\documentclass[journal]{IEEEtran}

\usepackage{cite}

\ifCLASSINFOpdf
  \usepackage{graphicx}
\else
\fi
\usepackage{amsmath}
\usepackage{algpseudocode}
\usepackage{algorithm}

\usepackage{booktabs}

\ifCLASSOPTIONcompsoc
 \usepackage[caption=false,font=normalsize,labelfont=sf,textfont=sf]{subfig}
\else
 \usepackage[caption=false,font=footnotesize]{subfig}
\fi
\usepackage{footnote}

\usepackage{url}
\usepackage[hidelinks]{hyperref}

\usepackage[subpreambles=true]{standalone}
\usepackage{import}
\usepackage{footmisc}

\begin{document}
%

\title{Baloo: A Large-Scale Hybrid Soft Robotic Torso for Whole-Arm Manipulation}


%
%
%

\author{Curtis C. Johnson, Andrew Clawson,
        Marc D. Killpack
\thanks{All authors are with the Robotics and Dynamics Laboratory at Brigham
Young University in Provo Utah, USA.}
\thanks{This work was supported by the National Science Foundation under Grant No. 1935312}}

\maketitle
\begin{abstract}
Soft robotic actuators can simplify the design of controllers when operating in contact-rich environments. Importantly, their passive compliance fundamentally alters contact mechanics by smoothing impacts and distributing forces over large areas. By integrating soft actuators, we can perform high-impact, dynamic, and contact-rich tasks that are challenging or impossible for traditional rigid robots. In order to explore the benefits of passive structural compliance and learn to utilize it effectively, we present a prototype robotic torso named Baloo. Baloo's hybrid soft-rigid design incorporates both adaptability from soft components and strength from rigid components with two meter-long, pneumatic robot arms mounted on a rigid torso. The hybrid design is capable of lifting end effector payloads of up to 19 kg, far exceeding many hybrid robot designs. Such payloads are competitive with similar-sized rigid robots, but with a much higher strength-to-weight ratio. Through 30 physical whole-body grasping experiments, we also demonstrate how a simple control strategy can generalize for effective lifting across six challenging objects with diverse shapes, sizes, and weights. A 100\% success rate across all objects--achieved with the simple control strategy--underscores the potential of our hybrid soft-rigid robot design for contact-rich, whole-body tasks.

\end{abstract}


%
\IEEEpeerreviewmaketitle

\section{Introduction}
%
%
%
%
\IEEEPARstart{R}{obotic} manipulation has been studied extensively for decades and remains a challenging problem--especially when facing `open-world' manipulation tasks with objects of varying shapes, sizes, and weights in vastly differing environments. Some objects are difficult to interact with using only an end effector. The typical solution to this is to simply use a larger and/or stronger robot. This challenge arises frequently when interacting with unknown, large, or heavy objects. In this work we advocate for a more general manipulation paradigm that encourages engaging the entire structure of the robot, instead of just an end effector. This approach, known as whole-arm or whole-body manipulation, has the potential to significantly expand the range and diversity of objects that can be manipulated in the real-world.

\begin{figure}[ht]
    \centering
    \includegraphics[width=0.75\columnwidth]{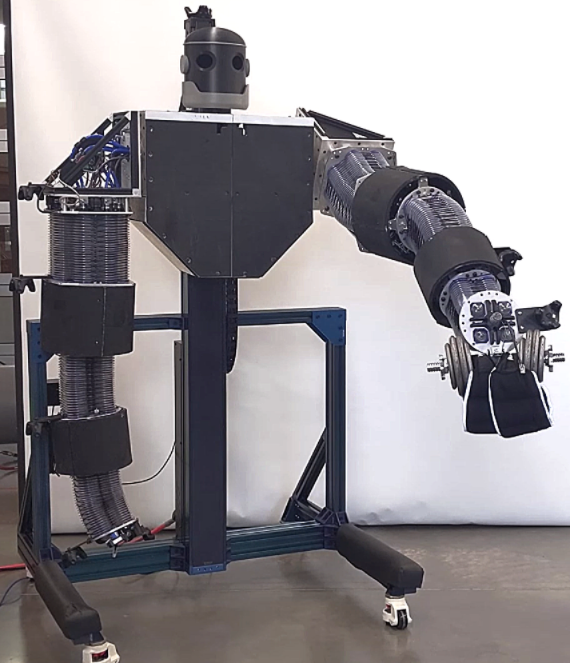}
    \caption{Photo of Baloo with one arm lifting a 19 kg weight. HTC Vive trackers are mounted on the end plate of each joint for joint configuration estimation. }
    \label{fig:baloo_glamor_shot}
\end{figure}

When operating in open-world environments, a successful manipulation system should at least be robust to uncertainty. Better still would be the ability to adapt to unknown objects and environments. To accomplish this, it is often useful for the robot to be compliant in some way. Compliance can be implemented on rigid robots with active joint-level control \cite{impedancectrl, Schumacher_Wojtusch_Beckerle_Von_Stryk_2019} and usually requires additional modeling, sensing, and high-bandwidth control loops. We suggest that the \textit{passive} compliance of soft materials offers some unique advantages over \textit{active} compliance, making it especially well-suited for contact-rich manipulation tasks in open-world environments.

One key advantage is how soft structures can change the nature of contact. Rigid contact is stiff, both physically and numerically. Soft contact has a smoothing effect and distributes sharp point contacts over larger surfaces. The ability to passively conform to unknown contact surfaces enhances grasp robustness. This property has been demonstrated in the design of soft grippers \cite{bubblegripper, chang2019alternative, shintake2018softgrippers}. Flexible elements also smooth out and absorb energy from unmodeled impacts that can otherwise damage rigid transmissions.

To take advantage of these benefits, one approach is to add compliant materials to an otherwise rigid robot. The authors of \cite{Salisbury_Townsend_Ebrman_DiPietro_1988} were among the first to recommend outfitting the Barret WAM (whole-arm manipulation) arm with high-friction, compliant material for whole-arm manipulation. Recent research supports this idea in several ways. The authors of \cite{Goncalves_Kuppuswamy_Beaulieu_Uttamchandani_Tsui_Alspach_2022, Zhang_Barreiros} use the rigid Kinova Jaco Gen2 robot arm outfitted with compliant pneumatic structures to accomplish whole-arm grasping of large objects. Other researchers have added compliant materials to the end effector only \cite{Zhang_Xie_Zhou_Wang_Zhang_2020} to accomplish tasks like bi-manual manipulation of a large box \cite{Dehio_Wang_Kheddar_2022}. Throughout this paper we refer to this type of robot as `rigid-soft', where the first word `rigid' defines the nature of the robot structure and actuation.

An alternative approach is to soften the structure and actuation of the robot itself. Because of the passive compliance built into the structure, these robots can excel in difficult tasks. Interesting examples include grasping awkward objects \cite{Becker_Teeple_Charles_Jung_Baum_Weaver_Mahadevan_Wood_2022}, contact-rich manipulation, human robot collaboration \cite{Jorgensen_Bojesen_Jochum_2022}, or tasks involving impacts\cite{Hughes_Culha_Giardina_Guenther_Rosendo_Iida_2016}. Soft robots offer several advantages over rigid robots in these situations, often with surprising robustness \cite{Bhatt_Sieler_Puhlmann_Brock_2021}. This robustness is possible because a soft structure performs a similar role as a low-level impedance controller, but acts with zero time delay in a decentralized manner without any additional sensors or actuators. Soft structures can also store potential energy for dynamic tasks like throwing \cite{Zwane2024} and exhibit stable oscillations, which is useful for locomotion \cite{Della_Santina_Duriez_Rus_2023}. Collectively, this idea of outsourcing control, sensing, and computational functionality onto hardware is known as mechanical intelligence \cite{Goncalves_Kuppuswamy_Beaulieu_Uttamchandani_Tsui_Alspach_2022,Hauser_Nanayakkara_Forni_2023}. We refer to this type of robot as `soft-rigid' to highlight that the actual structure of the robot is soft.

As many state-of-the-art soft robot platforms are designed on the millimeter to sub-meter scale, there are still significant challenges with scale and size \cite{Li_Awale_Bacher_Buchner_Della_Santina_Wood_Rus_2022}. The use of pneumatic actuation of soft bellows to address scaling issues is common in soft robotics \cite{Hashem_Stommel_Cheng_Xu_2021,Lamping_Muller_dePayrebrune_2022, Walker_Zidek_Harbel_Yoon_Strickland_Kumar_Shin_2020}. Recent research \cite{Hussain_Ficuciello_2024} suggests that incorporating both soft and rigid components into the structure of a robot can help solve problems that currently limit the utility of purely soft robot platforms. The relatively rigid components can carry heavier loads and improve force transmission, while the compliant components preserve passive compliance and adaptability \cite{Oh_Rodrigue, Yang_Asbeck_2020, Oh_Lee_Shin_Choi_Cho_Rodrigue_2024}. 

Despite significant advancements in soft robotics and the increasing use of passive compliance for contact-rich tasks, little to no prior work has demonstrated whole-body manipulation using soft or soft-rigid arms capable of handling significant payloads (6.75-21.9 kg). Existing designs focus instead on lightweight applications or manipulation tasks only involving the end effector. We help to bridge this gap for real-world tasks that demand both strength and flexibility.

In this work, we present our design of a soft-rigid robotic torso. A conceptually-similar version of the soft continuum actuator presented in this work is briefly mentioned in \cite{Felt_Telleria_Allen_Hein_Pompa_Albert_Remy_2019}, but the actual designs are not presented in detail. We discuss our modified design in detail, including several major improvements. In summary, the contributions of this paper are the following.

\begin{itemize}
    \item We present our design for a large-scale robot platform named Baloo (see Fig. \ref{fig:baloo_glamor_shot}), designed with a hybrid soft-rigid methodology to combine the strength of rigid components with the adaptability of soft components. 
    \item We highlight Baloo's unique hardware capabilities through several whole-body manipulation tasks using a variety of challenging objects that would be impossible for comparable state-of-the-art manipulators. Video\footnote{\label{video}\url{https://youtu.be/Aza158cfQSw}} and experimental data\footnote{\label{datacode}\url{https://github.com/byu-rad-lab/baloo-data-analysis}} are available online.
\end{itemize}

The remainder of the paper is structured as follows: Section \ref{sec:design} presents the design of the platform. Section \ref{sec:integration} describes details on the system integration, kinematics, and dynamics of Baloo. Section \ref{sec:experiments} outlines the hardware experiments used to test our design. We finish with a discussion of the results in \ref{sec:discussion} and conclude with \ref{sec:future}.

\section{Methods}
\subsection{Soft-rigid Robot Design}
\label{sec:design}
In this section we present design details for each subsystem on Baloo (Fig. \ref{fig:baloo_glamor_shot}). We start with the compliant joints, then discuss how the joints are assembled to build the pneumatic arms and how the arms operate in the context of the entire torso assembly. 

\subsubsection{Compliant Joints}
\label{sec:compliant_joints}

\begin{figure*}[tb]
    \centering
    \subfloat[]{\includegraphics[trim=0 50 0 50, clip, width=0.16\textwidth]{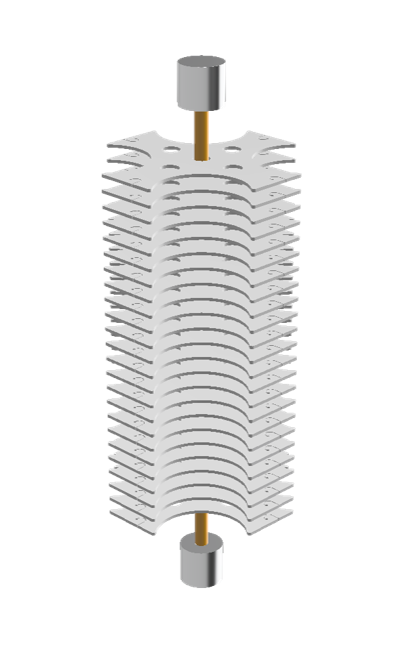} \label{fig:spine}  } \hfill
    \subfloat[]{\includegraphics[trim=0 50 0 50, clip, width=0.16\textwidth]{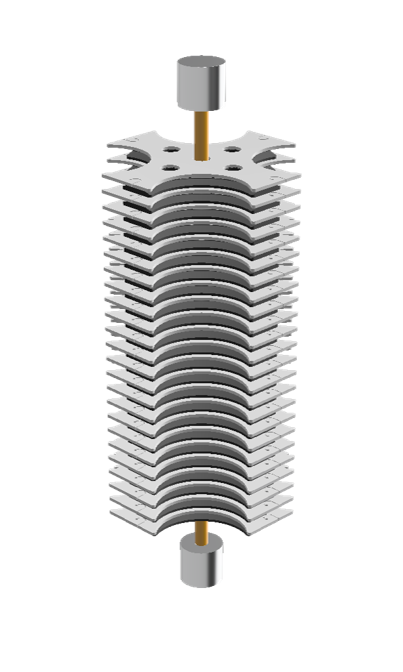} \label{fig:foam}} \hfill
    \subfloat[]{\includegraphics[trim=0 50 0 50, clip, width=0.16\textwidth]{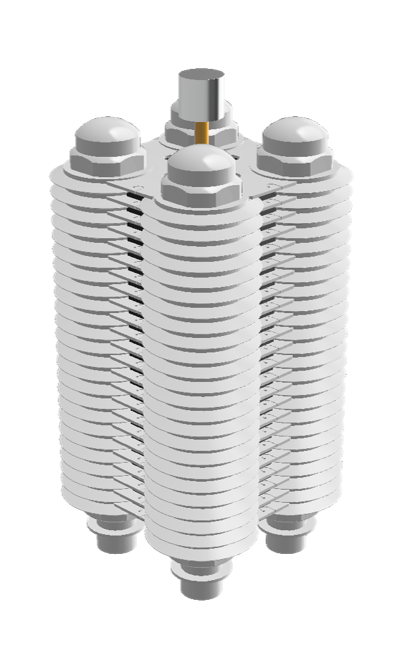} \label{fig:bellows}} \hfill
    \subfloat[]{\includegraphics[trim=0 50 0 50, clip, width=0.16\textwidth]{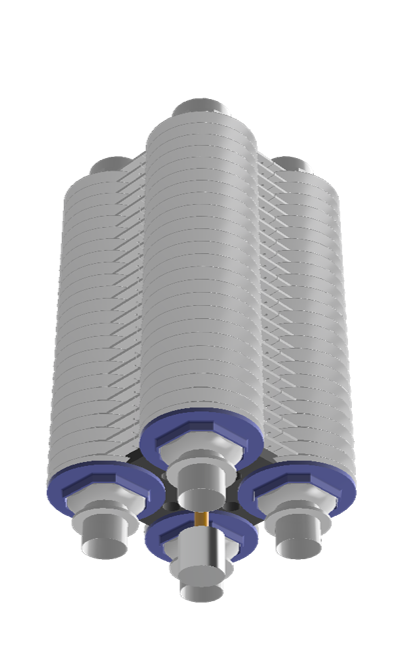} \label{fig:bottom_gaskets}} \hfill
    \subfloat[]{\includegraphics[trim=0 50 0 50, clip, width=0.16\textwidth]{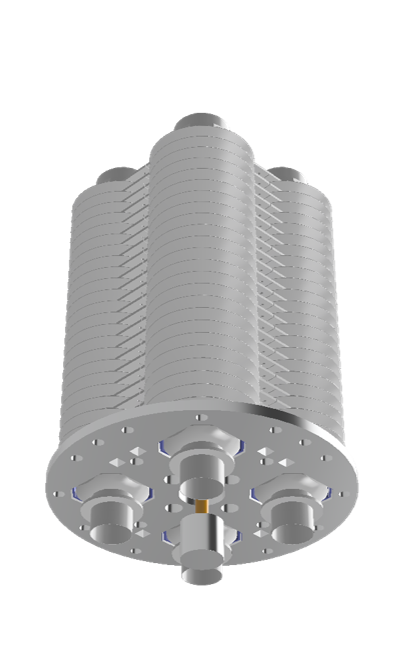}\label{fig:bottom_plate}} \hfill
    \subfloat[]{\includegraphics[trim=0 50 0 50, clip, width=0.16\textwidth]{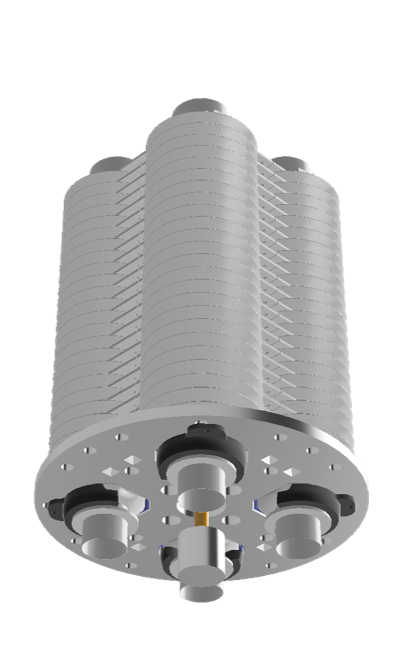}\label{fig:retaining_ring}} \\
    \subfloat[]{\includegraphics[trim=0 50 0 50, clip, width=0.16\textwidth]{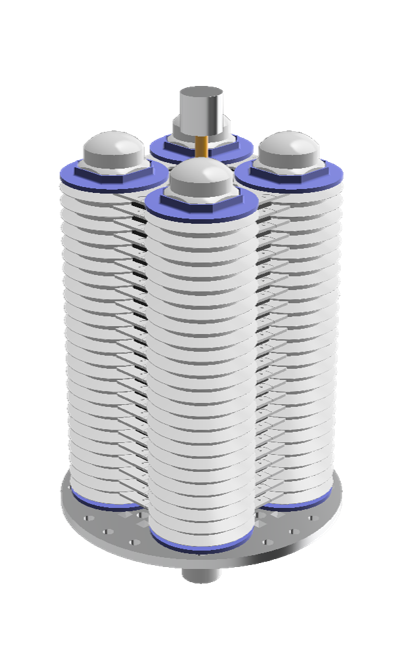}\label{fig:top_gasket}} \hfill
    \subfloat[]{\includegraphics[trim=0 50 0 50, clip, width=0.16\textwidth]{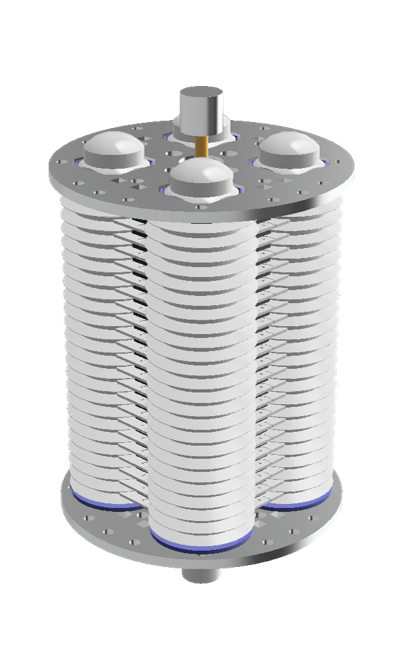}\label{fig:top_plate}} \hfill
    \subfloat[]{\includegraphics[trim=0 50 0 50, clip, width=0.16\textwidth]{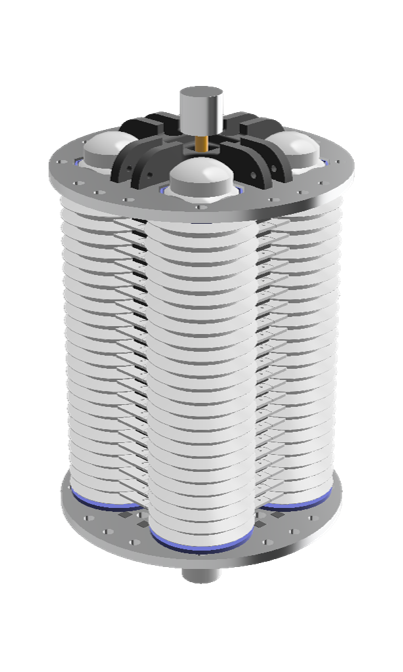}\label{fig:key_support}} \hfill
    \subfloat[]{\includegraphics[trim=0 50 0 50, clip, width=0.16\textwidth]{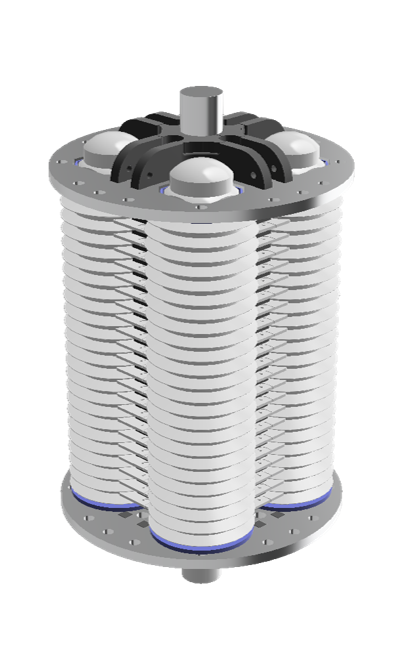}\label{fig:rope_key}} \hfill
    \subfloat[]{\includegraphics[trim=0 50 0 50, clip, width=0.16\textwidth]{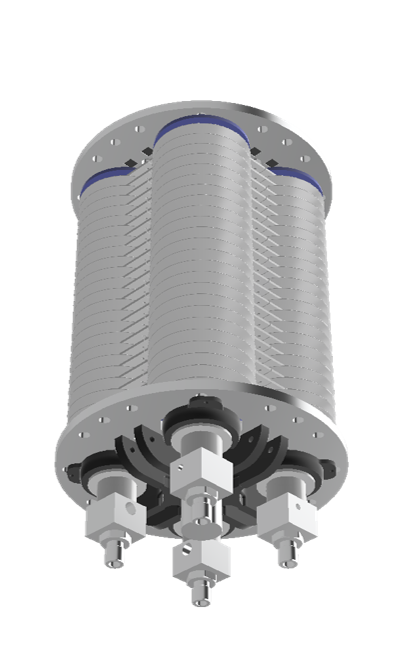}\label{fig:valve_adapter}} \hfill
    \subfloat[]{\includegraphics[trim=0 50 0 50, clip, width=0.16\textwidth]{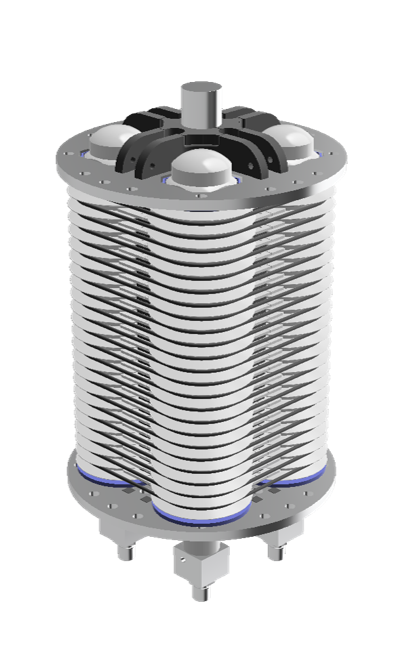}\label{fig:wire_rope}} \\

    \caption{Step-by-step assembly of a small compliant joint. The medium and large joints are assembled in a similar fashion.}
    \label{fig:lego_steps}
\end{figure*}

\begin{figure}[]
    \centering
    \includegraphics[width=\columnwidth]{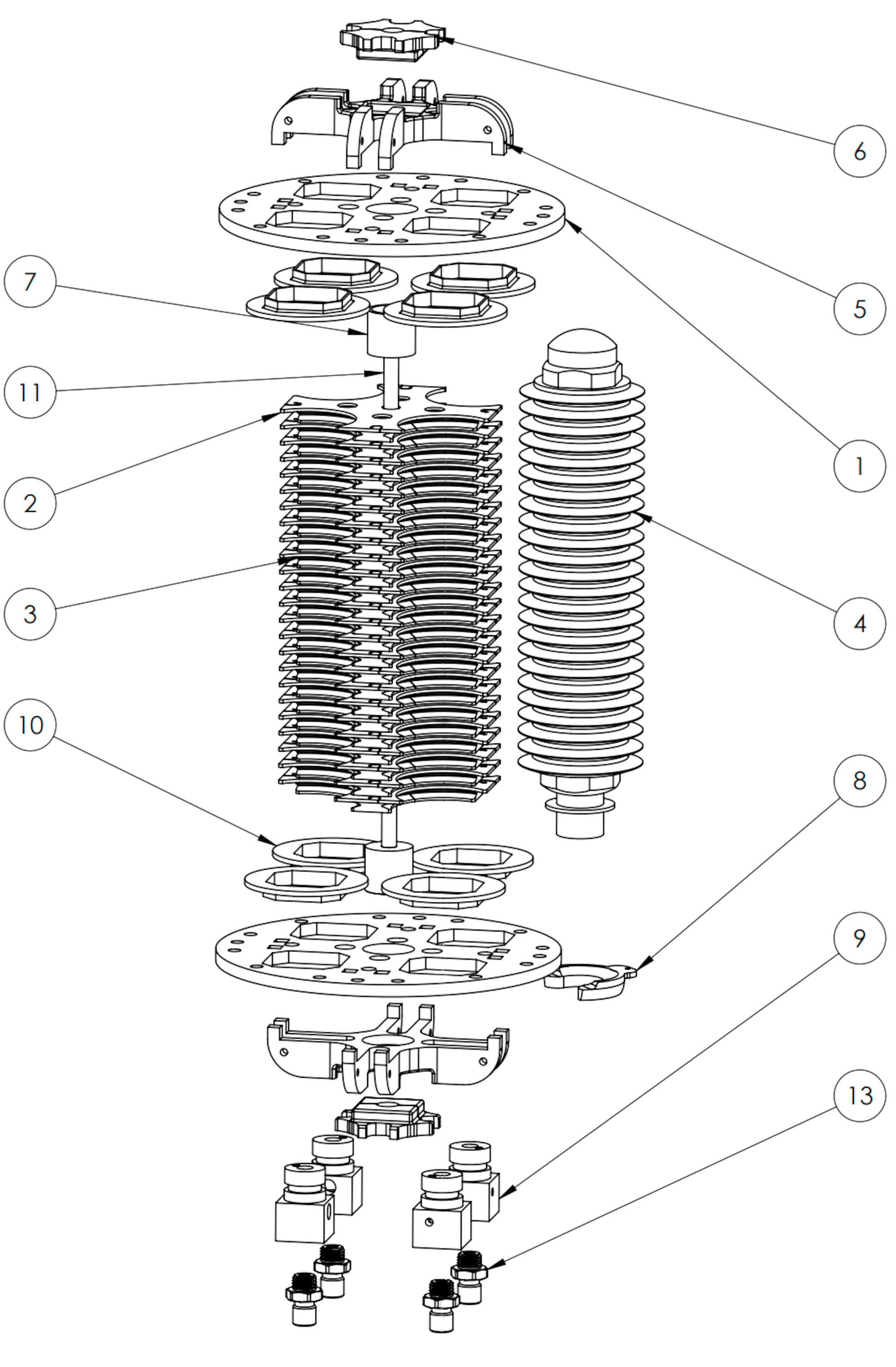}
    \caption{Exploded assembly view of a single small joint. Three additional plastic actuation chambers (\#4) are hidden to simplify the view.}
    \label{fig:exploded}
\end{figure}

This discussion will follow the step-by-step assembly process shown in Fig. \ref{fig:lego_steps}. For added clarity, specific components are also enumerated in Fig. \ref{fig:exploded}. 

\begin{figure}[]
    \centering
    \subfloat[]{\includegraphics[width=.4\columnwidth]{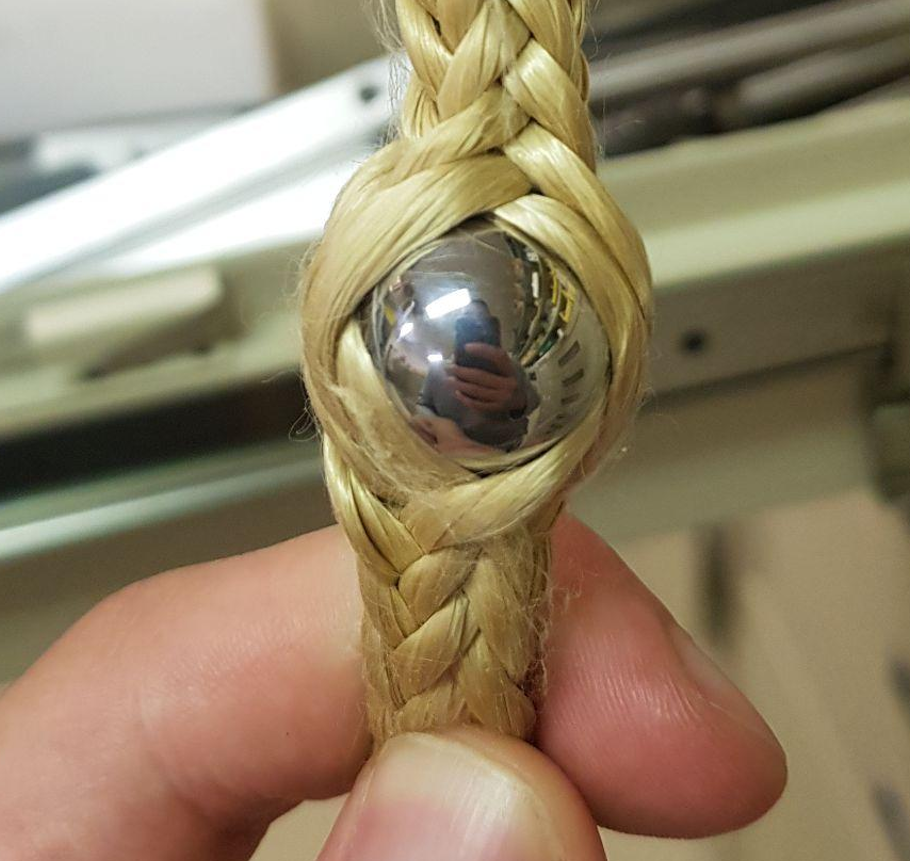}%
    \label{fig:steelball1}}
    \hfil
    \subfloat[]{\includegraphics[width=.43\columnwidth]{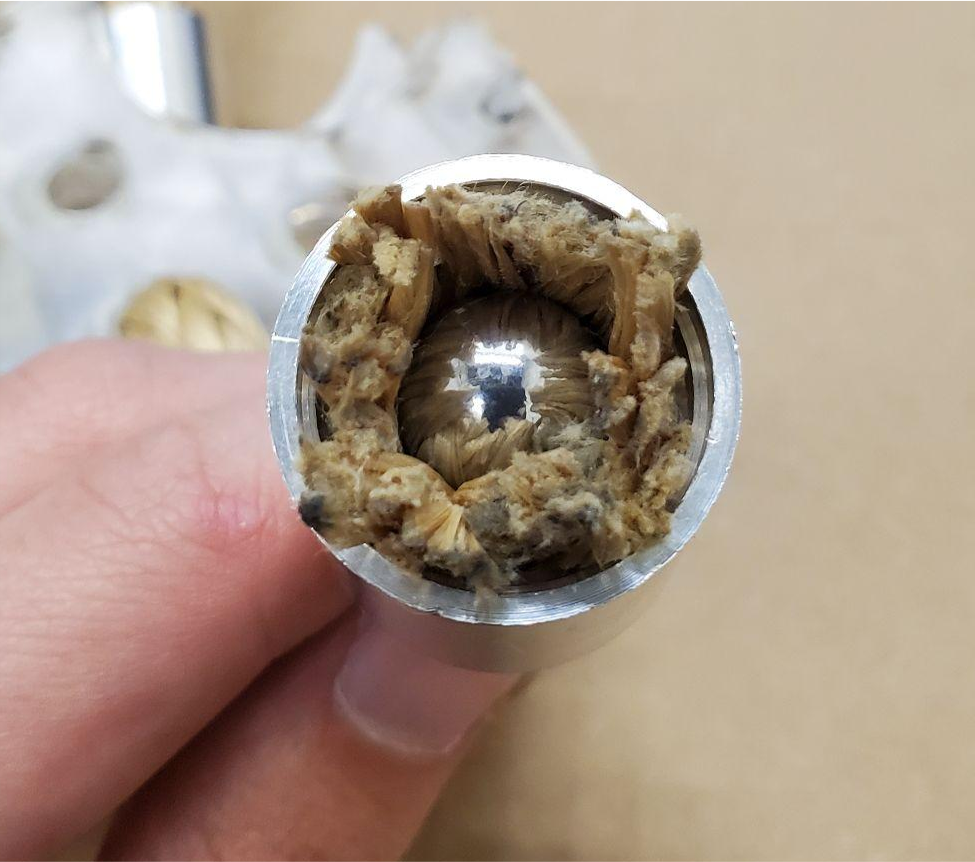}%
    \label{fig:steelball2}}
    \caption{Steel ball embedded in aramid rope braid and inserted into end fitting.}
    \label{fig:steel ball}
\end{figure}

Fig. \ref{fig:spine} illustrates the spine of the compliant joints, consisting of a series of Polyethylene terephthalate glycol (PET-G) spacers (Fig. \ref{fig:exploded}, \#2) stacked around an aramid rope (Fig. \ref{fig:exploded}, \#11). The spacers have several holes to facilitate internal routing of wires and air supply tubing. To constrain the joint’s length, an aluminum end fitting (Fig. \ref{fig:exploded}, \#7) is rigidly attached to each end of the aramid rope. The end fitting is a cylindrical bar with a conical hole at one end. A steel ball bearing, embedded within the rope weave (Fig. \ref{fig:steelball1}), is inserted into the conical hole of the end fitting (Fig. \ref{fig:steelball2}) and secured with epoxy to prevent the rope fibers from unraveling. Experimental testing showed that incorporating a steel ball increases the tensile strength of the spine assembly by approximately 50\% compared to epoxy alone. The rope length is slightly shorter than the resting length of the pneumatic chambers, ensuring that the fully assembled joint places the rope under tension.

Fig. \ref{fig:foam} shows the integration of open-cell polyethylene foam sheets (Fig. \ref{fig:exploded}, \#3) between each plastic spacer. The foam serves two key functions: 1) it maintains equal spacing between the spacers as the joint deflects and 2) it absorbs energy, acting as a mechanical damper within the joint.


To actuate the joint, we use a set of blow-molded Polyethylene Terephthalate (PET) pressure chambers featuring an accordion-like structure that enables bending, elongation, and compression (Fig. \ref{fig:exploded}, \#4). These chambers are arranged antagonistically in a ‘+’ pattern to allow for stiffness modulation. Each chamber is integrated into the spine, with spacers aligning with grooves in the accordion structure, as shown in Fig. \ref{fig:bellows}.

Once all four chambers are in place, they are fitted with protective TPU gaskets (Fig. \ref{fig:exploded}, \#10), seated into the holes of the bottom aluminum plate (Fig. \ref{fig:exploded}, \#1), and secured from underneath using a 3D-printed PLA retaining ring (Fig. \ref{fig:exploded}, \#8), as shown in Fig. \ref{fig:bottom_gaskets}, \ref{fig:bottom_plate}, and \ref{fig:retaining_ring}. The same procedure is applied to the top plate, except that no retaining ring is used (as the top of the plastic chamber lacks a groove for one) as illustrated in Fig. \ref{fig:top_gasket} and \ref{fig:top_plate}. Finally, a 3D-printed support structure (Fig. \ref{fig:exploded}, \#5) is inserted into the top and bottom plates, as shown in Fig. \ref{fig:key_support}.
    
Recall that the assembled structure shown in Fig. \ref{fig:key_support} will be slightly longer than the spine shown in Fig. \ref{fig:spine}. This is because the spine is intentionally designed to be slightly shorter than the resting length of the pressure chambers.

To constrain the overall length of the joint, a locking key, 3D-printed in PLA with 100\% infill (Fig. \ref{fig:exploded}, \#6), is inserted between the support structure (Fig. \ref{fig:exploded}, \#5) and the rope end fittings (Fig. \ref{fig:exploded}, \#7) at both ends of the joint. Installing the key at the bottom is straightforward since the aramid rope remains loose. However, inserting the key at the top (Fig. \ref{fig:rope_key}) requires compressing the assembly by pressing down on the top plate (Fig. \ref{fig:exploded}, \#1) until enough space is available to fit the key beneath the top end fitting. The result is an internal static equilibrium: the pressure chambers remain in compression, while the aramid rope is held in tension by the key assembly, with the tensile load distributed across most of the end plate via the support structure (Fig. \ref{fig:exploded}, \#5).

This tensioning design offers several key benefits. First, it allows for easy disassembly by simply compressing the joint and removing the key—an essential feature for replacing pressure chambers, swapping foam types, or changing broken spacers. Second, keeping the aramid rope in tension helps constrain the joint to a specific height, which is useful for kinematic modeling. Third, the tension ensures firm contact between the pressure chambers and end plates, preventing issues such as backlash and inefficient power transmission.

To complete the joint assembly, aluminum interface blocks (Fig. \ref{fig:exploded}, \#9) are inserted into the bottom of each pressure chamber and sealed with an O-ring, as shown in Fig. \ref{fig:valve_adapter}. These blocks provide access to the chambers for airflow through tube fittings (Fig. \ref{fig:exploded}, \#13) and allow for pressure sensing. Finally, as shown in Fig. \ref{fig:wire_rope}, the entire joint is wrapped with a plastic-coated steel cable that sits in the chamber grooves, preventing the pressure chambers from deforming radially away from the spine when pressurized.

\begin{figure}[]
    \centering
    \includegraphics[width=\columnwidth]{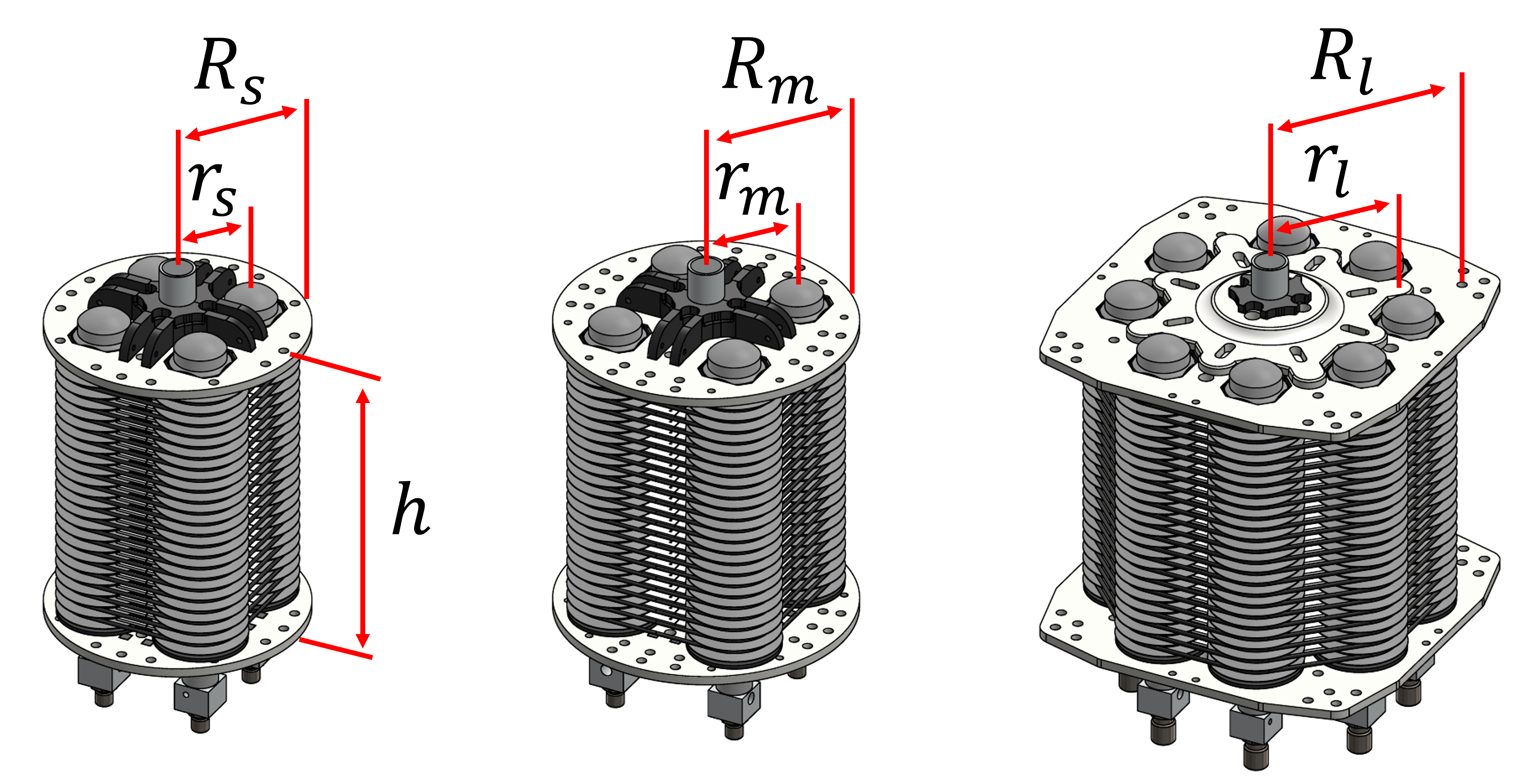}
    \caption{All three sizes of the passively compliant joints. Each size was selected based on torque requirements as discussed in Section \ref{sec:compliant_joints}. The subscripts $s$, $m$, $l$ refer to the small, medium, and large joint sizes listed in Table \ref{tab:joint_params}.}
    \label{fig:three_joints}
\end{figure}

While Fig. \ref{fig:lego_steps} illustrates the assembly steps for the small joint, each arm consists of a small, medium, and large joint, as shown in Fig. \ref{fig:three_joints}. All joints share a similar structure, including a central spine, pressure chambers, and end plates with locking keys. The primary difference among them is the joint radius. The guiding design principles for selecting these values are as follows:

\begin{enumerate}
\item Each pressure chamber has a maximum and minimum arc length $s_{max/min}$ (i.e., mechanical limits), which determines the maximum bending angle $\theta$ for a given radius $r$ using the relation: $s_{max} = r \theta$. Here, $r$ is the distance from the center of the spine to the center of a given pressure chamber.
\item The bending torque $\tau$ is related to $r$ by $\tau = r \times F$, where $F$ is the force exerted on an end plate by a pressurized chamber.
\item The joint mass increases with $r$.
\end{enumerate}

As $r$ increases, both torque output and mass increase, while the maximum bending angle $\theta$ decreases. To meet different performance requirements, we select different values for $r$, as listed in Table \ref{tab:joint_params}.

Since the torque requirements for the distal joint are the lowest, our nominal design starts with the small joint, positioning the pressure chambers as close together as possible (i.e., minimizing $r$). This results in the lowest mass, the largest bending range $\theta$, and the smallest torque output. The medium joint has an $r$ value 1.24 times that of the small joint, resulting in a proportional torque increase and a similar mass increase. While $r$ could be increased further, we prioritized reusing as many components from the small joint as possible. The potential torque benefits of a larger $r$ did not justify the additional weight and manufacturing complexity.

The large joint, responsible for carrying the full arm and any payload at the distal end of the arm, has the highest torque demands. Increasing $r$ alone to accommodate expected payloads is impractical, so we use four dual pressure chambers instead of four single chambers. This configuration doubles the torque output for a given $r$ while adding less mass, as the plastic pressure chambers are significantly lighter than the metal spine and plates. For the final large joint design, we increase $r$ to its practical limit (1.67 times the small joint) and double the torque output with additional chambers. This results in an effective torque increase of 3.33x (1.67 $\times$ 2) while keeping the mass increase to only 2.81x.

\begin{table}[]
\centering
\caption{Design parameters for each size of joint.}
\label{tab:joint_params}
\begin{tabular}{lllll}
\hline
Joint Size & $h$  & $R$ & $r$ & Mass \\ \hline
Small & 220 mm & 89 mm& 51 mm& 1.65 kg\\
Medium & 220 mm & 102 mm& 63 mm& 2.03 kg\\
Large & 220 mm & 128 mm& 85 mm& 4.63 kg\\ \hline
\end{tabular}
\end{table}

\subsubsection{Pneumatic Actuation}
\begin{figure}[]
    \centering
    \subfloat[]{\includegraphics[width=.5\columnwidth]{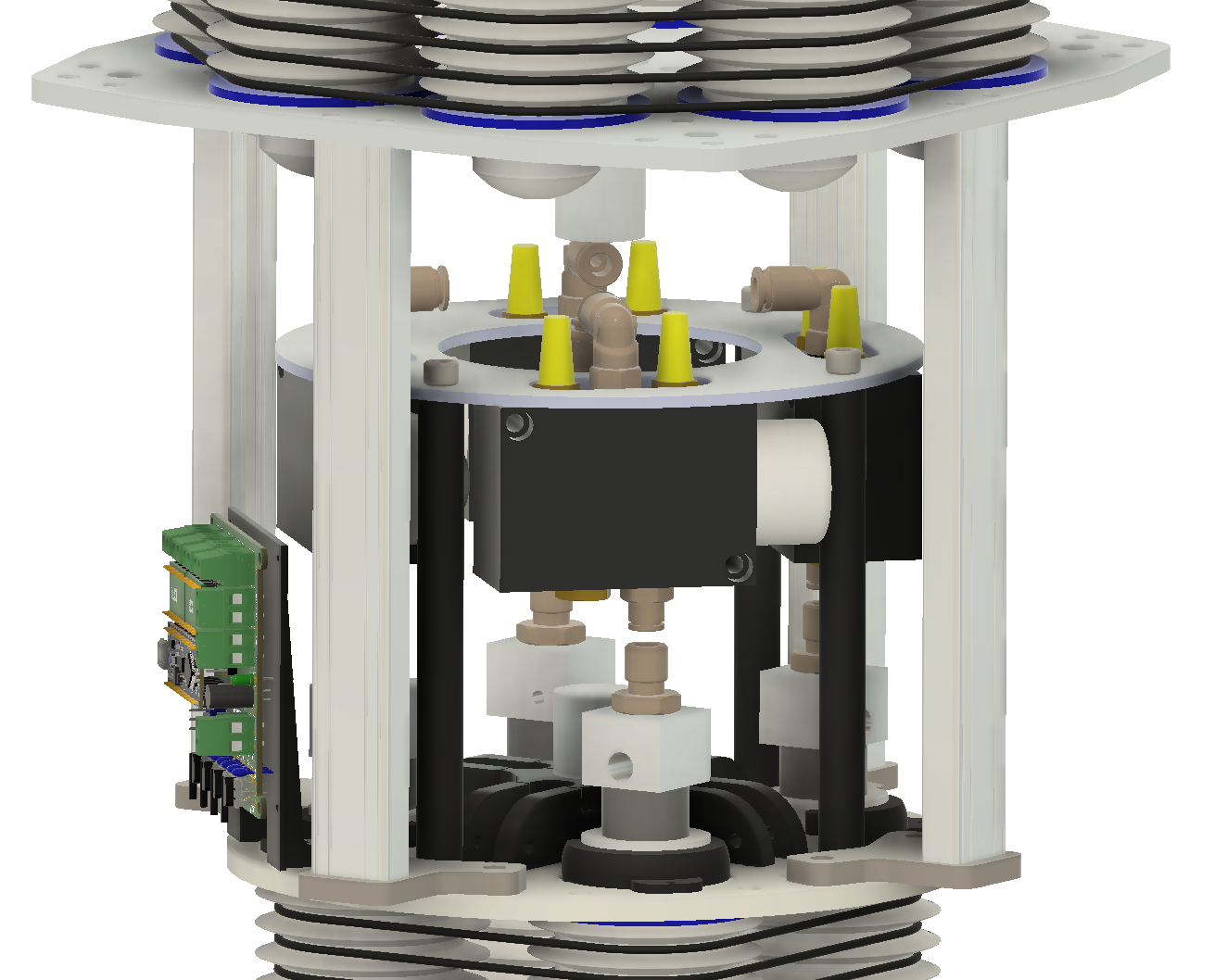}%
    \label{fig:link0}}
    \hfil
    \subfloat[]{\includegraphics[width=.35\columnwidth]{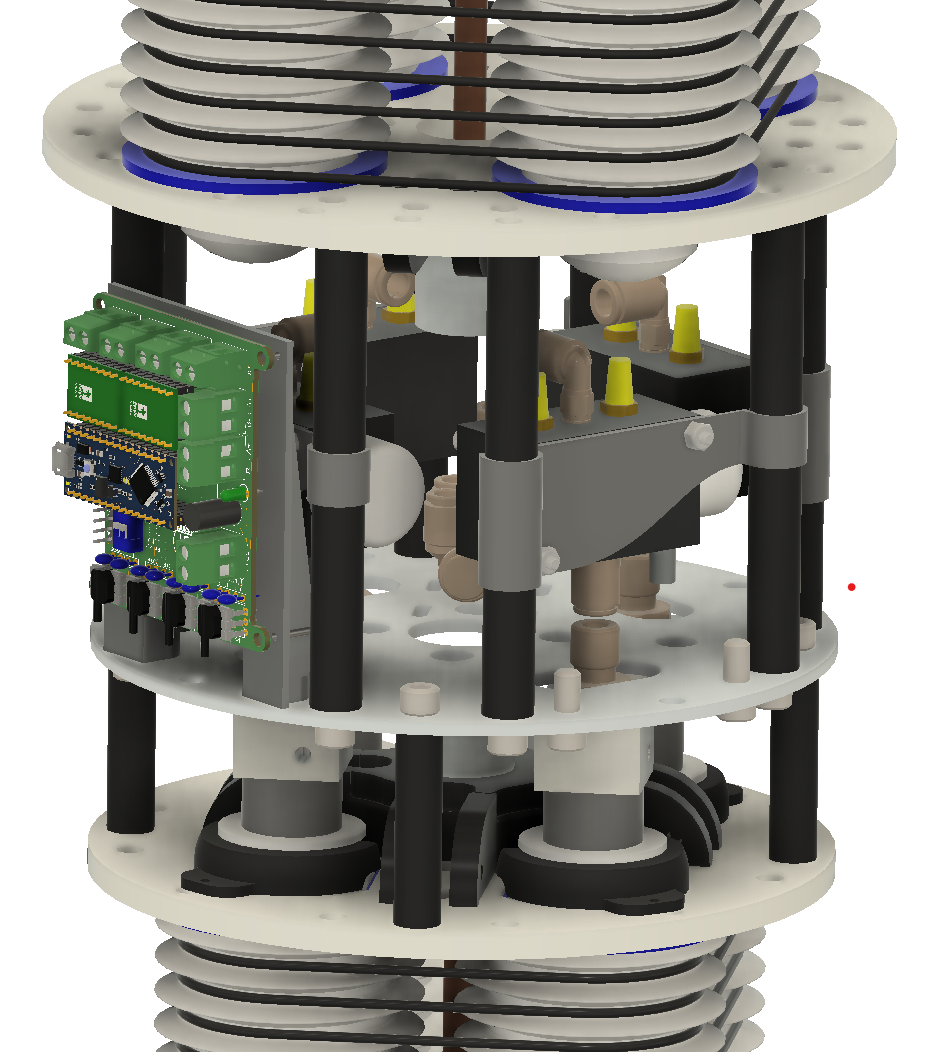}%
    \label{fig:link1}}
    \caption{a: Rigid link between large and medium joints, containing four valves which drive the chambers in the medium joint. (b): Rigid link between medium and small joints, containing four valves driving the chambers in the small joint.}
    \label{fig:links}
\end{figure}

Each of the compliant joints is actuated by individually filling or venting each of the pressure chambers. We use one Enfield LS 5/3 Proportional Directional valve to control a single chamber (or dual chamber for the large joints), for a total of 12 valves per arm. The valves are controlled with the PneuDrive pressure control system \cite{Johnson_Cheney_Cordon_Killpack_2024}. Between the large and medium joints is a rigid link containing four LS-V15s valves and associated control electronics for the medium joint, shown in Fig. \ref{fig:link0}. Between the medium and small joints is a similar rigid link, with four LS-V05s valves and electronics to control the small joint, shown in Fig. \ref{fig:link1}. The rigid links serve a dual purpose. First, they house the valve assembly near the joint. Placing the valves centrally would require at least 12 supply tubes and additional wiring to pass through the arm, creating space management challenges. By positioning the valve assemblies near the joint, both air and wiring can be organized as a central supply with drop points at each link, needing only two holes. In addition, placing the valves near the pressure chambers minimizes pressure losses in the tubing given the proximity of both the available source pressure and the incorporated pressure sensors. This helps to increase control bandwidth and improve responsiveness. Second, along with the rigid end plates of the joints, the links bear compressive and tensile loads, preventing excessive deformation of the compliant joints and improving force transmission.

Four LS-V25s valves and electronics to control the large joint are located in the chest, shown in Fig. \ref{fig:assembly}.

\subsubsection{Base and Torso}
\begin{figure}[]
    \centering
    \includegraphics[width=0.8\columnwidth]{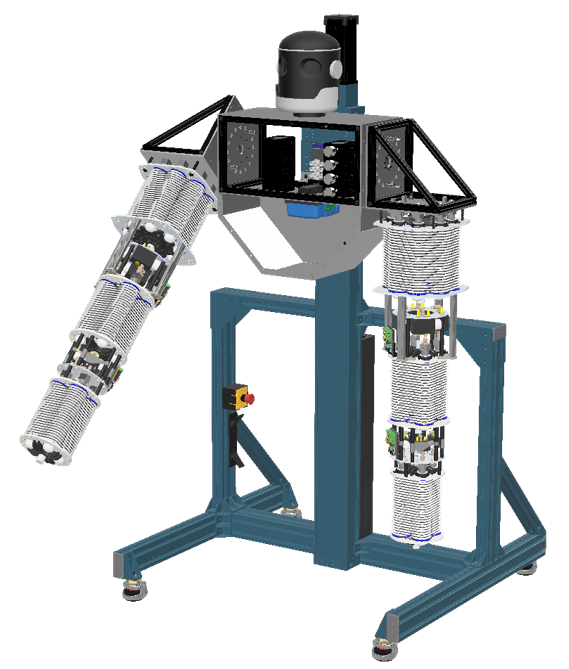}
    \caption{Assembly of the entire bi-manual robot with base, chest, and both arms. The right shoulder is rotated forwards to 45 degrees and the chest cover plates are removed for clarity.}
    \label{fig:assembly}
\end{figure}

Baloo is designed to have a workspace mostly in front of the robot, from the floor to approximately human height. Accordingly, the chest and arms are mounted on the Vention MO-LM-039-1530 Ball Screw Actuator, which is supported by a rigid structure of aluminum extrusion. The chest and arms are nominally at a height of 1.5 meters (as shown in Fig. \ref{fig:assembly}) and can be lowered all the way to the floor.   

The arms are mounted on the torso with ball bearings, which allow the arms to rotate outwards and be configured to reach in front of the robot. Each arm can be manually pinned into the desired configuration (i.e. a fixed angle relative to the chest) for manipulation tasks as shown in Fig. \ref{fig:assembly}.  


\subsection{Sensors, Kinematics, and Dynamics}
\label{sec:integration}
As previously mentioned, we use the modular pressure control system called PneuDrive \cite{Johnson_Cheney_Cordon_Killpack_2024} to control the four independent pressures for each joint. Each joint has an embedded pressure control module running high-rate PD control on pressures. The modules are connected to a serial communication bus running through the middle of the joints into the chest, where an Odroid N2+ acts as a controller for a single arm. Air, power, and communication lines travel down the arm using the holes in the spacers along the spine in Fig. \ref{fig:spine}. One of the Odroid computers also acts as controller for ball screw actuator that moves the torso up and down.

\begin{figure*}[tb]
    \centering
    \resizebox{\textwidth}{!}{
\begin{tikzpicture}
\node (input) [] {};
\node (controller) [block, right=1.5cm of input]{Joint Controller};
\node (pdcontroller) [block, right=1.5cm of controller] {$u = \mathrm{PD}(p_{des}, p)$};
\node(elevator_traj)[block, above=2cm of pdcontroller]{Ruckig};
\node(elevator_input)[left=1cm of elevator_traj]{};
\node(stepper)[block, right=1cm of elevator_traj]{Stepper Driver};
\node(elevator)[block, right=1.5cm of stepper]{Elevator};
\node(elevator_output)[right=2cm of elevator]{};
\node (elevator_feedback)[right=1cm of elevator]{};
\node (elevator_feedback2)[below=1cm of elevator_feedback]{};
\node (robotdynamics) [block, right=1cm of pdcontroller] {
$\begin{aligned}
    M(q) \ddot{q} + C(q, \dot{q}) \dot{q} + g(q) &= K_\tau p - d(q,\dot{q}) \\
    \dot{p} &= f(q,\dot{q}, p, u)
\end{aligned}$};
\node (right) [right=.5cm of robotdynamics] {};
\node (feedback) [below=1.cm of right]{};
\node (output) [right=1.cm of right] {};
\node [fit= (pdcontroller) (robotdynamics) (right) (feedback), draw, dashed, inner sep=0.25cm] (box){};
\node [below=0.1cm of box.south] {Pressure Controlled Robot Arm};

\node (stateoutput) [right=1cm of output]{};
\node (statefeedback) [left=1cm of stateoutput]{};
\node (statefeedback1) [below=2.5cm of statefeedback]{};

\draw[arrow] (input) -- (controller) node[midway, above]{$q_{cmd}$};
\draw [arrow] (controller) -- (pdcontroller) node[midway, above] {$p_{des}$};
\draw [arrow] (pdcontroller) -- (robotdynamics) node[midway, above] {$u$};
\draw [arrow] (robotdynamics.east) -- (right.center) |- (feedback.center) -| (pdcontroller.south) node[near end, left] {$p$};
\draw [arrow] (robotdynamics.east) -- (stateoutput.center) node[near end, above] {$q,\dot{q},p$};
\draw [arrow] (statefeedback.center) |- (statefeedback1.center) -| (controller.south);

\draw[arrow](elevator_input.center) -- (elevator_traj.west) node[midway, above]{$h_{cmd}$};

\draw[arrow](elevator_traj.east) -- (stepper.west) node[midway, above]{$h_{des}$};

\draw[arrow](stepper.east) -- (elevator.west) node[midway, above]{PWM};

\draw[arrow](elevator.east) -- (elevator_output.west) node[near end, above]{$h$};

\draw[arrow](elevator_feedback.center) --(elevator_feedback2.center) -| (elevator_traj.south);

\end{tikzpicture}
}
    \caption{Control Diagram of the Arms. We use the Ruckig Motion Planning \cite{berscheid2021jerk} library to generate smooth elevator motions. The dynamics of an arm are given by the mass matrix $M$, the Coriolis matrix $C$, the pressure-to-torque mapping matrix $K_\tau$, and the gravity torques $g$. $d$ is a vector of unknown/poorly modeled disturbances (e.g. stiffness, damping, etc). The pressure dynamics are listed explicitly here as some function $f$ of input and configuration variables.}
    \label{fig:control-diag}
\end{figure*}
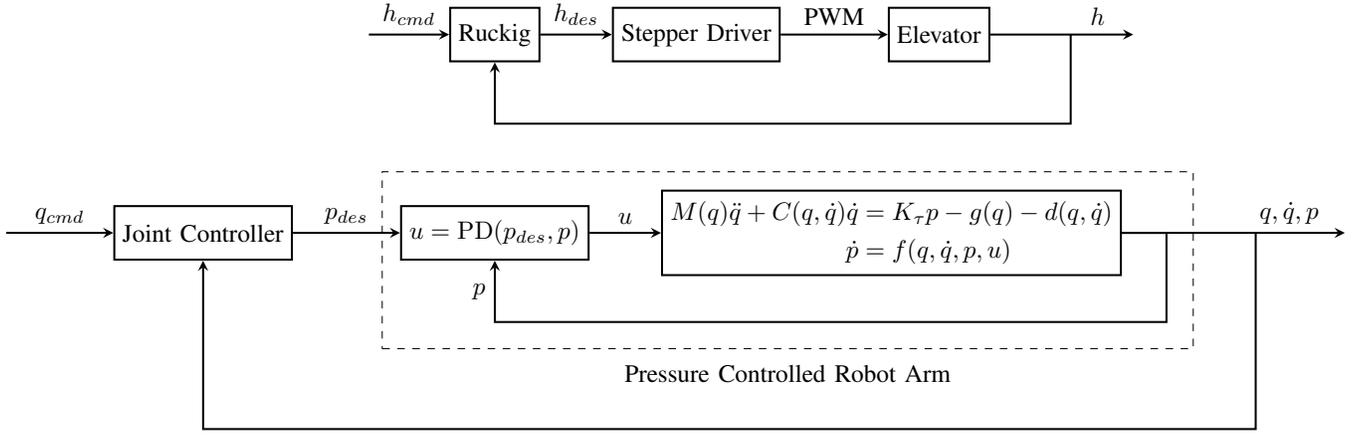

The Odroid N2+ computers are both connected via ethernet, which allows the entire system to be controlled via ROS. The control structure of the entire system is shown in Fig. \ref{fig:control-diag}.

Sensory feedback for a single arm includes pressures in each of the pressure chambers $p \in \mathcal{R}^{12}$, joint positions $q \in \mathcal{R}^6$, and joint velocities $\dot{q} \in \mathcal{R}^6$, as shown in Fig. \ref{fig:control-diag}. The height of the chest $h$ is another configuration variable. 

We use HTC VIVE Motion Trackers to track the orientation of both end plates of each joint (see the black trackers in Fig. \ref{fig:baloo_glamor_shot}) and calculate joint angles using the constant-curvature, singularity-free kinematic parameterization from \cite{Allen_Rupert_Duggan_Hein_Albert_2020, Hyatt_Kraus_Sherrod_Rupert_Day_Killpack_2019}. This parameterization assumes that the spine bends in a constant curvature arc and assigns two configuration variables $u$ and $v$ to the relative bending about the $x$ and $y$ joint axes respectively. Thus the joint position vector for a single arm is $q = [u_0, v_0, u_1, v_1, u_2, v_2]^T$.

The generalized torque vector for a single joint is given by the manipulator dynamics shown in Fig \ref{fig:control-diag}: 

\begin{equation}
    \tau = K_\tau p - d(q,\dot{q}) 
\end{equation}
\noindent where $K_\tau$ is the pressure-to-torque mapping matrix, $g$ is the gravity torques, and $d$ is a vector of unknown/poorly modeled disturbances (e.g. stiffness, damping, etc). 

The pressures $p$ are all measured and reported via PneuDrive. By having access to $p$ at any given instant we have an estimate of the actuation torques caused by the differential pressure between two antagonistic chambers ($K_\tau p$). Coupled with a model of the passive spring and damping torques of the compliant joint $d(\cdot)$, this low-level pressure feedback facilitates joint torque estimation. The pressures also respond to different loading conditions because the volume of the compliant chambers changes during deformation. This effect is captured by the pressure dynamics model

\begin{equation}
    \dot{p} = f(q,\dot{q}, p, u)
\end{equation}

\noindent where $f$ is some function that describes how pressures change in response to valve voltage commands $u$. In this work, we will not go into detail on specific choices for this model, but the authors of \cite{Johnson_Cheney_Cordon_Killpack_2024} evaluate several functional forms for it.

Joint position controllers can provide pressure commands $p_{des}$ given commanded joint angles $q_{cmd}$. To reduce undesirable mechanical vibrations for the elevator, we use the Ruckig \cite{berscheid2021jerk} Motion Planning library to plan a time-optimal jerk, acceleration, and velocity constrained trajectory $h_{des}$ in real time, given a commanded height $h_{cmd}$ (see Figure \ref{fig:control-diag}).

\subsection{Hardware Experiments}
\label{sec:experiments}

We perform three different experiments to validate our design. The first experiment examines joint-level dynamic characteristics. The second experiment tests the end-effector payload capability of a single arm. The third experiment consists of 5 sets of whole-body grasping trials with 6 different objects. 

\subsubsection{Dynamic System Identification}
\label{sec:systemid}

We performed two system identification tests to determine the dynamic characteristics of the compliant joints. The first test captured the free response of a single joint after an initial displacement. This test gives us a rough approximation of important dynamic parameters like stiffness and damping.

For the second test, we measured the mapping from pressure differential (i.e. torque) to joint angle. The pressure differential is the difference in pressure between antagonistic pressure chambers. This test moves through a full range of angles and highlights the nonlinearities present in the system. We gathered the data by commanding pressure differentials as a slow ramp and measuring the corresponding joint angle. 

\subsubsection{End Effector Payload}
\label{sec:payload}

To test the payload capabilities of our arm design, we rigidly attached a weight to the distal link of the arm, as shown in Fig. \ref{fig:baloo_glamor_shot}. The arm starts in a relaxed configuration with the weight attached to the end effector. We slowly increase the pressure differentials to their in each joint until they reach the maximum differential (i.e. 0 and 410 kPa). The failure pressure for these actuation chambers is theoretically 615 kPa but we enforce a regulated maximum pressure of 410 kPa to maintain a safety factor of 1.5. 

A standardized way of measuring maximum payload for soft robots does not exist in the literature. In an effort to compare against designs summarized in Table \ref{tab:literature_specs}, we define maximum payload as the maximum weight that can be lifted from a relaxed state to a height of 1 meter. The actual height is arbitrary, but we choose 1 meter since that is roughly the length of the arm. This implies that the arm keeps most of its range of motion under the load.

Because of the passive compliance of the arm, the joints naturally deflect in response to external loads. In essence, this is identical to the methods used in \cite{Bruder_Graule_Teeple_Wood_2023} in measuring the end effector height as a function of differential pressure.

\subsubsection{Whole-Body Grasping}
\label{sec:whole-body-grasp}
\begin{figure}[]
    \centering
    \includegraphics[width=\columnwidth]{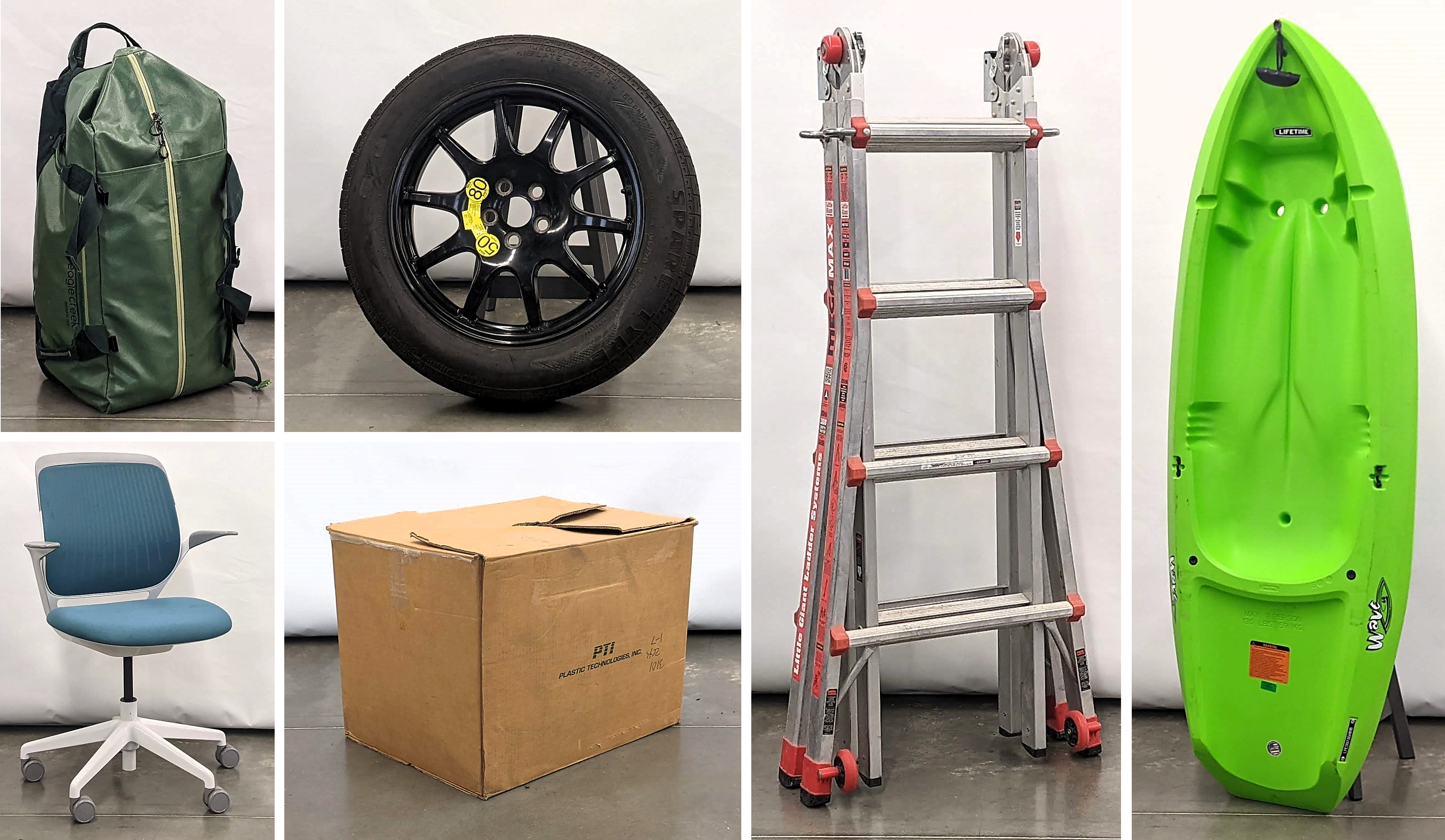}    
    \caption{Objects used to perform whole arm grasping trials}
    \label{fig:objects}
\end{figure}

To evaluate the effectiveness of the hybrid soft-rigid robot design in offloading some of the computational complexity of manipulation to the soft robotic hardware (i.e. mechanical intelligence), we perform whole-arm grasping experiments. We use several large, heavy, and unwieldy objects, listed in Table \ref{tab:whole-arm-trials} and pictured in Fig. \ref{fig:objects}. We chose these objects to span a variety of sizes, textures, weights, and shapes. No object-specific information is used in any part of the trials. Though we typically use the HTC Vive system for joint configuration estimation as discussed in Section \ref{sec:integration}, we do not use them for this experiment. The only feedback we use is the low-level pressures for closed loop pressure tracking.

\begin{table}[]
\centering
\caption{Manipuland descriptions }
\label{tab:whole-arm-trials}
\begin{tabular}{@{}llllll@{}}
\toprule
Item       & BBox (m)       & Mass (kg) & Challenges \\ \midrule
Car Tire  & 0.8 x 0.8 x 0.2  &  21.9     &   Large, Heavy      \\
Office Chair&0.7 x 0.7 x 1.0 & 11.9     &   Non-convex, Hinged       \\
Ladder & 1.2 x 0.6 x 0.2 & 15.6          &     Slippery, Heavy      \\
Duffel Bag & 0.4 x 0.4 x 0.8 &6.75        &     Deformable       \\
Kayak & 1.9 x 0.7 x 0.3 & 8.5               &     Large, Unwieldy      \\
Large Box  &0.6 x 0.5 x 0.5 & 7.3          &      Nonuniform mass dist.      \\
\bottomrule
\end{tabular}
\end{table}

As noted in Table \ref{tab:whole-arm-trials}, the box has a nonuniform mass distribution because it is filled with weights that are free to move around. The mass is initially concentrated at the bottom center of the box, but the weights move around as the box tilts during manipulation. This is also the case with the duffel bag, which is filled with foam holding a single weight.

The hardware experiment follows Algorithm \ref{alg:grasping_experiment}. We start with a manually designed trajectory $P$ that linearly interpolates from a nominal gravity compensation set point $p_{grav}$ to a closed grasp. Note that the pressure trajectory $P$ ranges from 0 to 300 kPa. We chose 300 kPa as a maximum to be conservative and avoid crushing objects.

There are 24 pressures (i.e. 12 for each arm) and $N$ total steps, each sent at 20 Hz. We run five trials for each of the six objects. After the arms lift up to compensate for gravity, we command the elevator to move to the centroid of the object. Once the elevator arrives at the commanded location, we place the object in the graspable workspace in front of the torso and between the arms. We made no attempt to do this repeatably so there are natural variations in position and orientation between each trial. These variations can be seen in the video\footref{video}. We classify the outcome of each trial into three categories: drop, slip, or lift. A drop indicates that the attempt failed completely. A slip indicates that the object may have slipped out of the grasp, but did not fall to the ground. A lift indicates a successful grasp.

\begin{algorithm}[]
    \caption{Dual-Arm Grasping Experiment} \label{alg:grasping_experiment}
    \begin{algorithmic}[1]
    \State \textbf{Input:} Predefined pressure trajectory: $P \in \mathcal{R}^{24 \times N}$
        \For{each object (6 objects)}
            \State \textbf{Input:} Object centroid height $h_{centroid}$
            \For{each trial (5 trials per object)}
                \State Raise arms: $p_{cmd} \gets p_{grav}$
                \State Set elevator command: $h_{cmd} \gets h_{centroid}$
                \State Wait until $h = h_{cmd}$
                \State Place object in workspace \label{alg:object_placement}
                \For{$i=0$ to $N-1$} 
                    \State $p_{cmd} \gets P[:, i]$
                    \State Wait for .05 seconds
                \EndFor
                \State Command elevator to lift: $h_{cmd} \gets 0$
                \State outcome $\gets$ "Drop" or "Slip" or "Lift"
                \State Reset: $p_{cmd} \gets 0$
            \EndFor
        \EndFor
    \end{algorithmic}
\end{algorithm}

\section{Results and Discussion}
\label{sec:discussion}

This section discusses the results from each of the three hardware experiments described in Section \ref{sec:experiments}.

\subsection{Dynamic System Identification}
\begin{figure}
    \centering
    \includegraphics[width=\columnwidth]{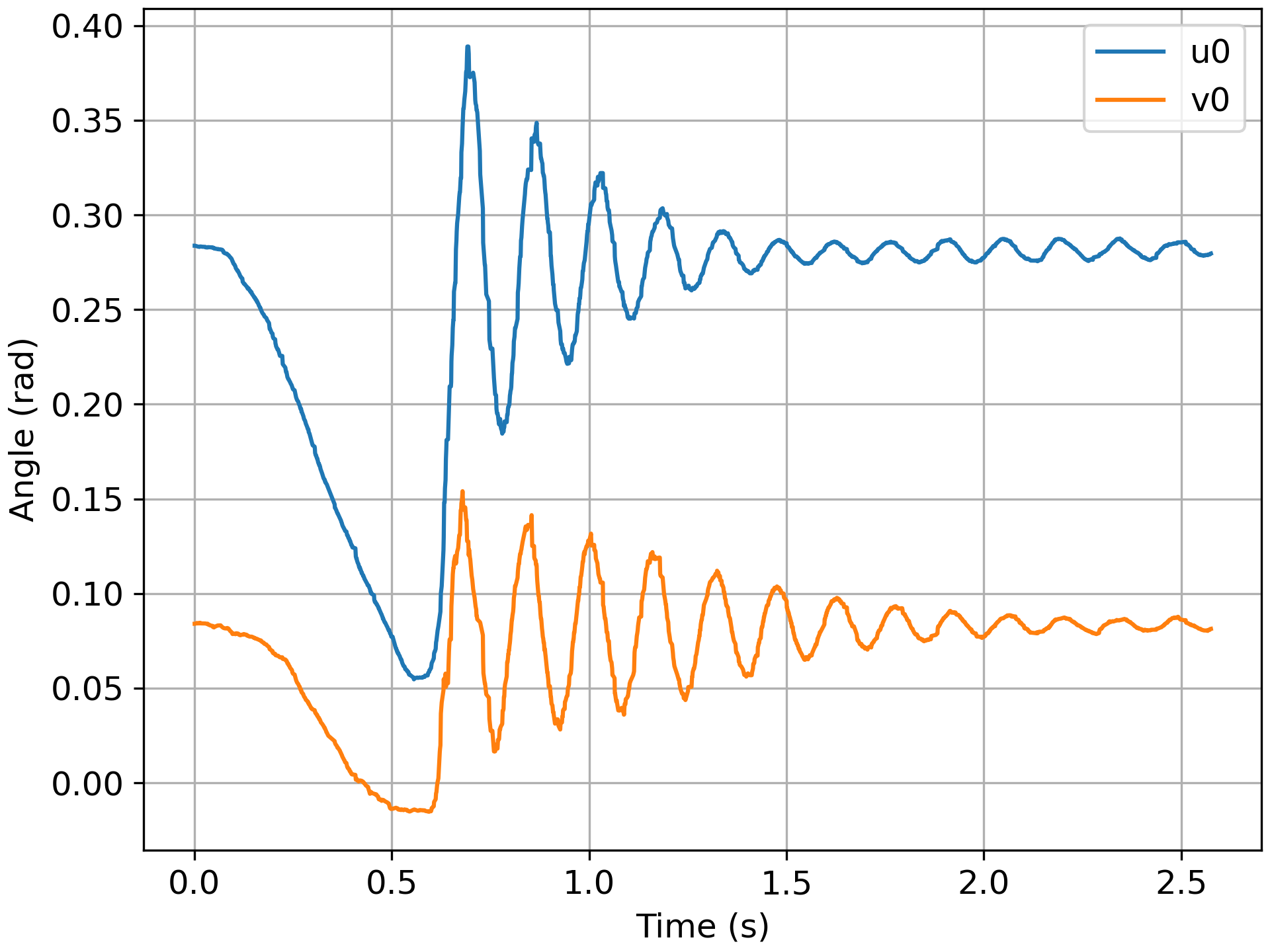}
    \caption{Free response of a single small joint along each degree of freedom. The logarithmic decrement method counting times at successive peaks gives an estimated damping ratio of .0015 and a damped natural frequency of about 8 Hz.}
    \label{fig:imp-resp}
\end{figure}

 It is well known that the dynamics of soft robot actuators are typically nonlinear with respect to the state ($q$, $\dot{q}$, $p$), and our design is no exception. However, in order to quickly gain intuition for the dynamic characteristics of the joints, we fit a linear second-order model to the free response shown in Fig. \ref{fig:imp-resp}. This is justified because the joint only deviates a small amount from an equilibrium position during the test. 

A linear second-order system is fully parameterized by the natural frequency $\omega_n$ and the damping ratio $\zeta$. We use the logarithmic decrement method \cite{Palm_2014} to estimate the damping ratio of the uncontrolled joint for both degrees of freedom and found that it is approximately $\zeta = .0015$. The period of oscillation is approximately 0.125 seconds, which means that the damped natural frequency $\omega_d = 8$ Hz. For a second order system, the natural frequency is related to the damped natural frequency by $\omega_d = \omega_n \sqrt{1 - \zeta^2}$, which gives a natural frequency of approximately 8 Hz as well. The highly underdamped nature of the joint is somewhat undesirable as it makes control difficult with pneumatics, which have a limited closed-loop control bandwidth. Future work will include improving the mechanical damping properties of the foam inside the joint to increase the damping ratio and keep parasitic torques low. Recall from Section \ref{sec:compliant_joints} that our joint design makes this a straightforward task, as the entire joint can be easily disassembled to swap foam spacers. 

\begin{figure}
    \centering
    \includegraphics[width=\columnwidth]{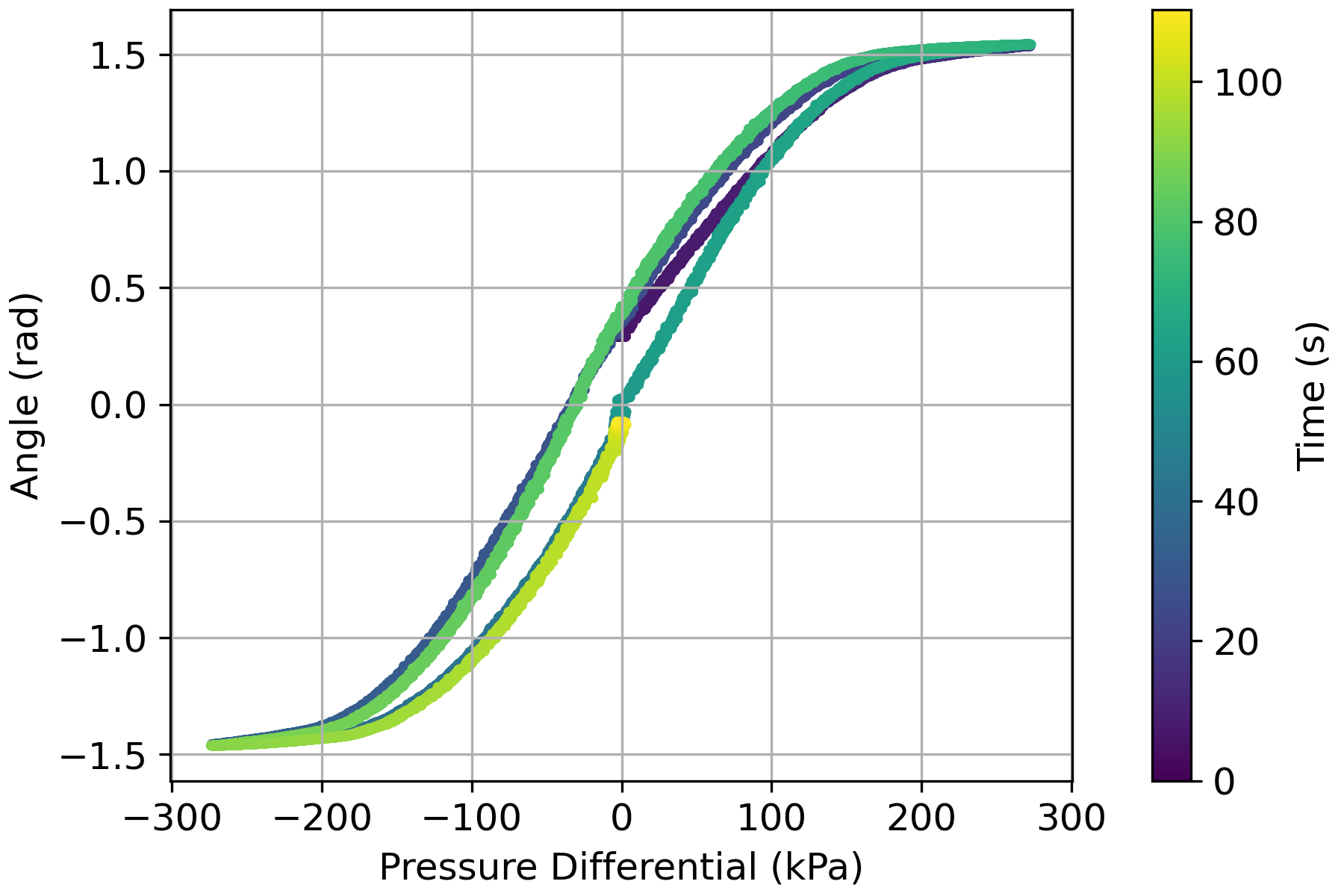}
    \caption{Typical hysteresis loop of a single small joint over two cycles. The pressure differential is the difference in pressures between antagonistic pressure chambers and is the quantity which causes a bending torque.}
    \label{fig:hysteresis}
\end{figure}

The results of the differential pressure to angle mapping are shown in Fig. \ref{fig:hysteresis}. It is clear from that there is a significant hysteresis loop present along each degree of freedom. We also observed that if the plastic chambers are deformed for long periods of time the position of the hysteresis loop shifts (i.e. the equilibrium position of the passive spring elements changes). 

\subsection{End Effector Payload}

An image of Baloo successfully lifting 19.3 kg off of the floor is shown in Fig. \ref{fig:baloo_glamor_shot}. A video showing lifting of consecutively higher weights is available online\footref{video}.

Shear deformation is particularly noticeable in the medium joint as the arm is extended straight forwards. This is typical and expected of continuum soft robot actuators under large loads.

For comparison, Table \ref{tab:literature_specs} summarizes some specifications of current state-of-the-art hybrid manipulators, along with some commercially-available rigid `cobot' manipulator specifications for additional context. It is important to note that the specifications listed in Table \ref{tab:literature_specs} are only for a single arm, with the assumption that any of these arms could also be used in a dual configuration. The term `rigid-soft' implies that an otherwise rigid robot is outfitted with compliant materials. A `soft-rigid' structure on the other hand, means that the structure and actuation of the robot is soft with added rigid components. As demonstrated in Figure \ref{fig:baloo_glamor_shot}, our design is capable of lifting end effector payloads of up to 19 kg which exceeds the maximum payloads of all comparable hybrid designs by at least 3.8x (and several commercially available rigid cobots). While some of the hybrid designs have a higher payload-to-mass ratio (2.5 for \cite{Liu_Shi_Chen_Yu_Zhang_Wang_2023} and 1.34 for \cite{Oh_Rodrigue}-a) than our design (1.28), no other hybrid manipulator is currently capable of performing the large-scale manipulation tasks demonstrated in Figure \ref{fig:6 objects}. It is unclear how state-of-the-art hybrid designs would scale to meet our payload requirements. For instance, material properties, actuator efficiency, and structural integrity may introduce diminishing returns or even negative trade-offs as size and load increase.

The payload values listed in Table \ref{tab:literature_specs} are for end-effector loads in a fully extended configuration, which is not the case for the whole-body grasping experiments. When the arms are not fully extended and the mass is located close to the chest, we expect the maximum lifting load to be limited by torque of the elevator stepper motor instead of the by the arms. We have only tested this motor up to a total lifting payload of 41 kg (20.5 kg/arm), but the theoretical maximum \textit{static} payload is roughly 120 kg--assuming no structural failures.

\subsection{Whole-Body Grasping}
The results for the whole body grasping trials are summarized in Table \ref{tab:experiment_results}. Every trial is shown side-by-side in the video\footref{video}. Every object was successfully lifted off the floor. The ladder was particularly challenging because of its low surface friction. It was never dropped, but slipped slightly as the object was lifted. The slippage stopped as the arms deformed and came into contact with additional areas along the ladder. This behavior highlights the benefit of a hybrid soft-rigid design. The soft elements automatically adapt to the slip and change in pose with zero-delay and no dedicated slip sensing. 

\begin{table}[]
\centering
\caption{Whole-Arm Grasping Results}
\label{tab:experiment_results}
\begin{tabular}{@{}llllll@{}}
\toprule
Item       & Drops      & Slips & Lifts \\ \midrule
Car Tire  & 0 &  0     &  5      \\
Office Chair& 0 & 0     &   5       \\
Ladder & 0 & 2          &     3      \\
Duffel Bag &0 &0        &     5      \\
Kayak & 0 & 0               &     5      \\
Large Box  &0 & 0          & 5       \\
\bottomrule
\end{tabular}
\end{table}

\begin{figure*}[tb]
    \centering
    \includegraphics[width=\textwidth]{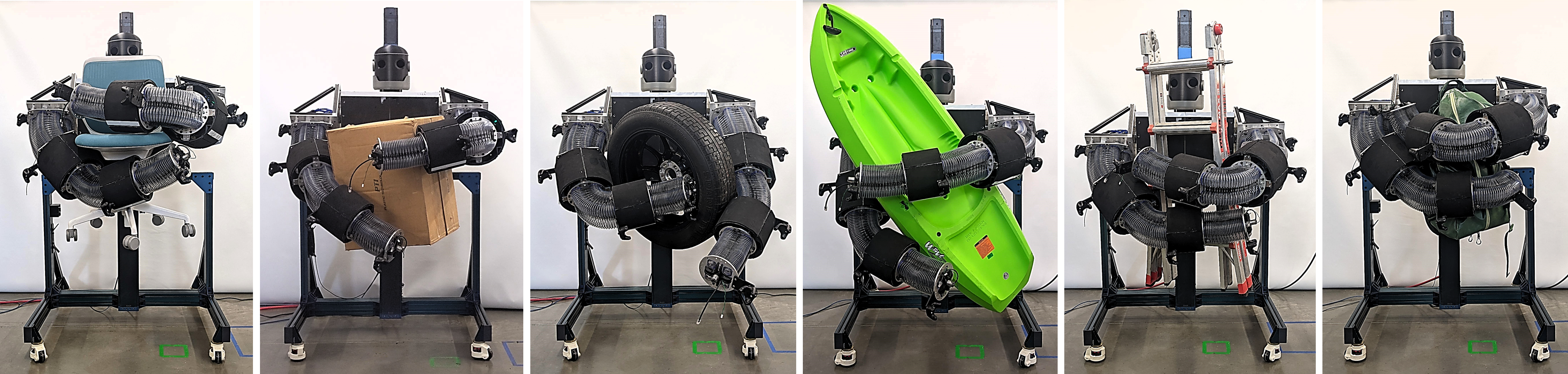}
    \caption{All six objects successfully picked up using the same pressure trajectory. The final grasp pose of the arms vary significantly, which demonstrates the adaptability of passive compliance.}
    \label{fig:6 objects}
\end{figure*}

Recall that we placed each object in the reachable work space in front of Baloo but did not put any constraints on the initial pose of the object--only its approximate placement. The accompanying video\footref{video} shows the variability in the initial placement of each object. This results in significant changes in the final grasping pose of the arms as shown in Fig. \ref{fig:6 objects}. The exact same pressure trajectory resulted in each of the different grasp poses shown. To the best of our knowledge, no other robotic manipulator (rigid or soft) has successfully manipulated such a wide variety of objects using only open-loop trajectories. This highlights the inherent adaptability of the hardware platform to conform to variations in size, geometry, and mass distribution while performing whole-body manipulation. It also illustrates the promise of whole-body manipulation with a large-scale soft-rigid robots.

\begin{table}[]
\label{tab:spec-comparison}
\centering
\caption{Comparison of current state-of-the-art manipulator capabilities of varying structural designs. Specifications are for a single-arm configuration. (NR - Not Reported.)}
\label{tab:literature_specs}
\begin{tabular}{@{}llllll@{}}
\toprule
Paper & Structure & Lift Payload (kg) & Mass (kg) & Length (m) \\ \midrule
  Panda  & Rigid & 3 & 18 & .85 \\
  UR10e  & Rigid & 12.5 & 33.5 & 1.3 \\
  LBR iiwa 14  & Rigid & 14 & 32.7 & .82 \\
  \cite{Goncalves_Kuppuswamy_Beaulieu_Uttamchandani_Tsui_Alspach_2022}  & Rigid-Soft & 2.6 & 4.4 & .985 \\
  \cite{Oh_Rodrigue}-a    &  Soft-Rigid        & 3         & 2.24   &   .85  \\
  \cite{Oh_Rodrigue}-b   &  Soft-Rigid       &  .624      & 16.85  & 5 \\
  \cite{Yang_Asbeck_2020} & Soft-Rigid & 1.3 & 1.6 & NR \\
  \cite{Oh_Lee_Shin_Choi_Cho_Rodrigue_2024}  & Soft-Rigid & 2 & 2.67 & .9 \\
  \cite{Bruder_Graule_Teeple_Wood_2023} & Soft-Rigid & 0.3 &  0.8 & .64\\
  \cite{Liu_Shi_Chen_Yu_Zhang_Wang_2023} & Soft-Rigid & 5 &  2 & .7\\
  \cite{Su_Qiu_Chen_Huang_Guan_Zhu_2023} & Soft-Rigid & 1 &  .642 & .36 \\
  This work            &  Soft-Rigid       &  19.3        & 15.1    & 1.1 \\ 
\end{tabular}
\end{table}

It is important to note some limitations of the whole-body grasping experiments. While we vary the object's initial pose during the experiments (Algorithm \ref{alg:grasping_experiment}, Line \ref{alg:object_placement}), we do ensure that the object could reasonably be grasped using the open-loop trajectory $P$. For example if the tire, kayak, or ladder were placed flat on the ground so that its major axis was parallel to the floor instead of perpendicular to it, the grasp would certainly fail. Our current grasping policy is open-loop and not reactive to the object pose, so during the object placement, we choose the nominal pose of the object such that its bounding-box lies within the anticipated workspace given $P$. This is not to say that Baloo is incapable of manipulating objects outside of this region, only that a simple open-loop `hugging' trajectory is not sufficiently intelligent to do so. Creating policies capable of tipping, bracing, and reorienting objects is an exciting direction for future work. The focus of this paper is to demonstrate the potential of hybrid soft-rigid designs for this type of manipulation, so the simple open-loop trajectory $P$ is sufficient for our objective.

\section{Conclusion and Future Work}
\label{sec:future}
This paper presents our prototype design of a bi-manual soft robot, built to explore the hybrid design philosophy of combining highly compliant structures with rigid ones. Our design enables whole-arm manipulation tasks that are impossible to accomplish using current state-of-the-art hybrid manipulator designs. Though the task could theoretically be accomplished with some rigid manipulators (e.g. the LBR iiwa 14 or the UR10e in Table \ref{tab:spec-comparison}), we demonstrate that our design simplifies control and improves adaptability. The simple grasping trajectory we use in this work would likely fail more often on these robots because stabilizing contact would be limited to fewer, smaller locations on the object. A hybrid soft-rigid robot design offloads a significant portion of the control complexity to the hardware and stabilizes contact by deforming around the object. 

Future work towards improving the performance of this prototype can include an in-depth investigation of specific material properties (e.g. stiffness and damping) to embed into the compliant joints to optimize its dynamic behavior. Refining the manufacturing process can also help to reduce the uncertainty we observed in material properties as well as kinematics. 

We also depend on the HTC Vive motion tracking system for configuration estimation. This is not ideal, especially for whole-arm manipulation tasks, as occlusions and accidental contact with trackers is likely. Future work will include embedding length sensors as shown in \cite{Allen_Rupert_Duggan_Hein_Albert_2020, ChristianPCC} into the joints to eliminate the need for an external motion capture system and measure non-constant curvature loading cases. 

Additionally, future research on outfitting Baloo with tactile sensing skin to close the loop during whole-body manipulation tasks should enable more advanced and complex bi-manual and whole-body manipulation. We expect that using fabric-based tactile sensing in a closed loop \cite{day2018scalable}, as opposed to the open-loop trajectories used in this work, will be essential to enabling improved manipulation capabilities. We also anticipate that the addition of onboard cameras could be used for both proprioception (e.g.  observing passive arm deformations) as well as exteroception (e.g. object pose estimation or grasp planning). 

Implementing control loops or learned policies using these additional sensing modalities could open the door to a wide variety of open-world manipulation capabilities that have so far been impossible.

\section*{Acknowledgment}

The authors would like to thank Christian Sorensen, Haley Sanders, Mihai Stanciu, Jake Sutton, Jon Black, Ryan Hall, Kendall Green, Kyler Nordgran, Mark Watson, Elton Harrison, and the engineering team at Vention for their help in designing, manufacturing, and assembling Baloo.

This work was supported by the National Science Foundation under Grant No. 1935312.

\section*{Conflict of Interest}
The authors declare no conflict of interest. 

\ifCLASSOPTIONcaptionsoff
  \newpage
\fi



\bibliographystyle{IEEEtran}
\bibliography{main.bib}

\end{document}